\documentclass{article}

\usepackage{microtype}
\usepackage{graphicx}
\usepackage{subfigure}
\usepackage{booktabs}
\usepackage{hyperref}
\usepackage{amsmath, amssymb, amsthm}
\usepackage{multirow}
\usepackage{makecell}
\usepackage[preprint]{icml2026} 
\usepackage{caption}
\usepackage{algorithm}
\usepackage{algorithmic}
\usepackage{multicol}
\usepackage{float}
\newtheorem{theorem}{Theorem}
\newtheorem{definition}{Definition}

\icmltitlerunning{Ca-MCF: Category-level Multi-label Causal Feature selection}

\begin{document}

\twocolumn[
\icmltitle{Ca-MCF: Category-level Multi-label Causal Feature selection}

\begin{icmlauthorlist}
\icmlauthor{Wanfu Gao}{jlu}   
\icmlauthor{Yanan Wang}{jlu}   
\icmlauthor{Yonghao Li}{swufe}
\end{icmlauthorlist}

\icmlaffiliation{jlu}{College of Computer Science and Technology, Jilin University, Changchun, China}
\icmlaffiliation{swufe}{School of Computing and Artificial Intelligence, Southwestern University of Finance and Economics, Chengdu, China}

\icmlcorrespondingauthor{Yonghao Li}{liyonghao@swufe.edu.cn}

\vskip 0.3in
]

\printAffiliationsAndNotice{}

\begin{abstract}
Multi-label causal feature selection has attracted extensive attention in recent years. However, current methods primarily operate at the label level, treating each label variable as a monolithic entity and overlooking the fine-grained causal mechanisms unique to individual categories. To address this, we propose a \underline{Ca}tegory-level \underline{M}ulti-label \underline{C}ausal \underline{F}eature selection method named Ca-MCF. Ca-MCF utilizes label category flattening to decompose label variables into specific category nodes, enabling precise modeling of causal structures within the label space. Furthermore, we introduce an explanatory competition-based category-aware recovery mechanism that leverages the proposed Specific Category-Specific Mutual Information (SCSMI) and Distinct Category-Specific Mutual Information (DCSMI) to salvage causal features obscured by label correlations. The method also incorporates structural symmetry checks and cross-dimensional redundancy removal to ensure the robustness and compactness of the identified Markov Blankets. Extensive experiments across seven real-world datasets demonstrate that Ca-MCF significantly outperforms state-of-the-art benchmarks, achieving superior predictive accuracy with reduced feature dimensionality.
\end{abstract}

\section{Introduction}
Feature selection has emerged as a fundamental technique to combat the curse of dimensionality by identifying a compact subset of relevant features \citep{guyon2003, jiang2022semi, li2023high, kamalov2023feature, ni2024feature}. By removing redundant and irrelevant attributes, this process significantly enhances model interpretability, reduces computational overhead, and improves generalization performance \citep{ling2019bamb, guo2022error, ling2024causal}. Multi-label feature selection (MLFS) extends these principles to more complex data structures where each instance is simultaneously associated with multiple labels, a paradigm that has become pervasive in domains such as image annotation and bioinformatics \citep{zhang2013, gibaja2015, wang2020towards, yu2021multilabel, li2023ssfs}. Modern MLFS methods have further evolved to address practical challenges such as missing features via implicit label replenishment \citep{xu2025missing}. Effective MLFS must not only identify features relevant to individual targets but also account for the intricate dependencies and correlations within the label space \citep{zhang2007, xu2016, wu2020mbmcf}. To this end, recent breakthroughs have integrated causal discovery \citep{chu2023continual, ling2024efficient} for optimal spouse identification and leveraged neural feature attribution for label-specific selection, maintaining robust predictive power in high-dimensional settings.

To achieve greater robustness and structural stability, research has increasingly shifted toward multi-label causal feature selection (MLCFS) \citep{huang2018manifold, zhang2019manifold, hu2022feature, zhang2023mfsjmi, li2023robust, faraji2024multilabel, dai2024multilabel, ma2025a, ma2025b}. This field focuses on identifying the Markov Blanket (MB) \citep{tsamardinos2003algorithms, guyon2007causal, jiao2024survey} of the label set, which represents the minimal set of features—comprising parents, children, and spouses—that renders the labels independent of all other variables \citep{yu2021}. By identifying this causal skeleton, MLCFS provides a more stable feature subset compared to traditional correlation-based methods.

Despite these developments, current MLCFS methods typically operate at the label level, treating each label variable as a monolithic entity. Early methods such as JFSC \citep{huang2018jfsc} and MCLS \citep{huang2018manifold} focus on joint learning and manifold-based constraints, followed by manifold regularized discriminative selection in MDFS \citep{zhang2019manifold}. The field then transitions toward local causal structure learning with methods like MB-MCF \citep{wu2020mbmcf} and M2LC \citep{yu2021}, which effectively identify Markov Blankets but treat features and labels as equally weighted nodes. More recently, Wu et al. \citep{wu2023practical} address the label blocking effect, wherein a dominant category masks the causal signal of a true feature, leading to its premature exclusion. Even with the latest advancements in label-aware selection such as LaCFS \citep{ling2025}, or the high-efficiency discovery processes of MI-MCF \citep{ma2025b} and MCF-Spouse \citep{ma2025a}, these methods maintain a coarse-grained perspective that overlooks the fine-grained causal mechanisms unique to individual categories within a label.

To address these limitations, we propose the Category-level Multi-label Causal Feature selection (Ca-MCF) method. By shifting the granularity of discovery from the label level to the category level through label flattening, Ca-MCF enables the precise modeling of causal structures that are otherwise obscured. Our main contributions are summarized as follows:
\begin{itemize}
    \item We introduce the concept of label-category flattening, a technique that decomposes multi-category labels into specific category nodes to facilitate granular causal analysis at the sub-label level.
    \item Two novel information-theoretic metrics, namely Specific Category-Specific Mutual Information (SCSMI) and Distinct Category-Specific Mutual Information (DCSMI), are proposed to measure correlations within the flattened category space and quantify inter-label category dependencies.
    \item An explanatory competition-based recovery mechanism is designed to restore valid causal features that are statistically suppressed by label-category interactions, thereby ensuring the completeness of the identified causal skeleton.
    \item The implementation of structural symmetry checks and cross-dimensional redundancy removal further refines the feature subset, ensuring the robustness and compactness of the identified Markov Blankets.
\end{itemize}

\section{Related Work}
Existing MLFS methods can be broadly categorized into four groups based on their underlying mechanisms: information-theoretic, regularization-based, manifold learning-based, and Bayesian network (BN) structure learning-based methods \citep{pereira2018categorizing, gonzalez2020distributed, wu2020mbmcf, yu2021, ma2025b}.

\textbf{Information-Theoretic Methods.} \citep{lee2015d2f} proposes the D2F method, which utilizes interaction information to measure dependencies among multiple variables; \citep{lee2015fast} develops the FIMF method, which improves time efficiency through an information-theoretic feature ranking mechanism; \citep{gonzalez2020distributed} proposes ENM to select the features with the largest \textit{$L_2 $}-norm; \cite{covert23a} develops an amortized optimization approach to greedily maximize conditional mutual information for dynamic feature selection;\citep{ma2025b} presents MI-MCF, which prioritizes the distinct contributions of labels and features by utilizing mutual information to optimize the search process.

\textbf{Regularization-based Methods.} \citep{ma2012web} proposes SFUS for automatic image annotation; \citep{faraji2024multilabel} develops the MLFS-GLOCAL method, which utilizes global relevance and local discriminative analysis to capture complex label relationships; \citep{huang2018jfsc} proposes the JFSC model, which integrates feature selection and classification into a unified regularization method. It selects task-relevant features and mines the shared feature subspace via sparse models, boosting annotation efficiency and accuracy through sparse constraints. \citep{li2024multi} presents the ESRFS method, which leverages an elastic net-based method for highly sparse and low-redundancy feature selection.

\textbf{Manifold Learning-based Methods.} \citep{cai2018multi} proposes MSSL, which incorporates feature manifold learning and sparse regularization to preserve the geometric structure of the data; \citep{huang2018manifold} utilizes MCLS via Laplacian scores to handle complex multi-label data; \citep{zhang2019manifold} develops the MDFS method, which applies manifold regularization to construct low-dimensional embeddings for discriminative selection.

\textbf{Causal and BN Structure Learning-based Methods.} Beyond correlation-based methods, causal discovery methods focus on identifying the MB to ensure structural stability and interpretability. \citep{wu2020mbmcf} proposes MB-MCF, which identifies the MB for multiple labels by learning local causal structures in a unified method. \citep{yu2021} develops M2LC, which formulates feature selection as a local causal discovery task and employs error-correction routines to handle dependencies within the label space. Furthermore, \citep{wu2023practical} proposes KMB, a robust method that discovers Markov boundaries without relying on the strong faithfulness assumption, effectively restoring causal signals masked by dominant associations. Despite these advancements, existing MLCFS methods predominantly operate at the monolithic label level\citep{yu2020causality}, treating each label as an indivisible entity. While the recently proposed LaCFS \citep{ling2025} explores category-specific causal relationships, it is designed exclusively for single-label multi-class scenarios and lacks the capacity to model the complex category-level interactions inherent in multi-label data. To the best of our knowledge, the proposed method is the first to address the fine-grained causal discovery problem in a multi-label context, unlocking the potential for more precise and granular multi-label causal analysis.

\begin{algorithm*}[t]
\caption{Ca-MCF: Category-level Multi-label Causal Feature selection}
\label{alg:master}
\begin{algorithmic}[1]
\REQUIRE Data matrix $\mathcal{D}$, labels $\mathcal{L}$, thresholds $\delta_1, \delta_2$, parameters $k_1, k_2$
\ENSURE Globally selected feature subset $selfea$
\STATE Initialize $global\_selected\_features \leftarrow \emptyset$
\FOR{each label $L_i \in \mathcal{L}$}
    \STATE $\mathcal{C}_i \leftarrow$ Get unique class values of $L_i$
    \FOR{each target class $c \in \mathcal{C}_i$}
        \STATE $target\_vec \leftarrow (L_i == c)$ \COMMENT{Construct class-specific binary vector}
        \STATE \textbf{Phase 1:} $Matrix\_R_{ic} \leftarrow$ Label-aware Dependency Modeling (Alg. 2)
        \STATE \textbf{Phase 2:} $[PC, SP] \leftarrow$ Local Structure Discovery (Alg. 3)
        \STATE \textbf{Phase 3:} $CMB \leftarrow$ Label-aware Feature Recovery (Alg. 4)
        \STATE \textbf{Phase 4:} $CMB \leftarrow$ Symmetry Check and Redundancy Removal (Alg. 5)
        \STATE $global\_selected\_features \leftarrow global\_selected\_features \cup CMB$
    \ENDFOR
\ENDFOR
\STATE \textbf{return} $selfea = \text{unique}(global\_selected\_features)$
\end{algorithmic}
\end{algorithm*}

\section{Preliminary}
\label{sec:preliminary}

In this section, we formally present the theoretical method and core definitions that underpin the Ca-MCF method. By shifting the focus from monolithic labels to fine-grained categories, we establish the necessary metrics and structural components for label-aware causal discovery.

\begin{definition}[Label-Category Flattening]
\label{def:flattening}
Let the multi-label set be $\mathcal{L} = \{L_1, L_2, \dots, L_L\}$, where each label $L_i$ contains $k_i$ distinct categories. The label-category flattening set is defined as $\mathcal{C} = \bigcup_{i=1}^L \{C_{i1}, C_{i2}, \dots, C_{ik_i}\}$. For each category $C_{ic}$, a binary vector $\mathbf{y}_{ic} \in \{0,1\}^N$ is constructed to represent the specific state of a label for $N$ samples, enabling category-level causal modeling.
\end{definition}

\begin{definition}[Class-Specific Causal Metrics]
\label{def:metrics}
To capture dependencies within the flattened space $\mathcal{C}$, we define two primary metrics in both unconditional and conditional forms:

\par \textbf{Specific Category-Specific Mutual Information (SCSMI)}: Quantifies the causal dependency between a feature $X$ and a target category $C_{ic}$. The unconditional SCSMI is defined as $SCSMI(X; C_{ic}) = I(X; C_{ic})$. The conditional SCSMI given a set $S$ is defined as:
\begin{equation}
\begin{split}
    SCSMI(X; C_{ic} \mid S) = H(X \mid S) - H(X \mid C_{ic}, S) \\
    = \sum_{x, c, s} p(x, c, s) \log \frac{p(x, c \mid s)}{p(x \mid s) p(c \mid s)}
\end{split}
\end{equation}

\par \textbf{Distinct Category-Specific Mutual Information (DCSMI)}: Quantifies the association between two label categories $C_{ic}$ and $C_{jd}$. The unconditional DCSMI is defined as $DCSMI(C_{ic}; C_{jd}) = I(C_{ic}; C_{jd})$. The conditional DCSMI given a set $S$ is defined as:
\begin{equation}
\begin{split}
    DCSMI(C_{ic}; C_{jd} \mid S) = H(C_{ic} \mid S) - H(C_{ic} \mid C_{jd}, S) \\
    = \sum_{c, y, s} p(c, y, s) \log \frac{p(c, y \mid s)}{p(c \mid s) p(y \mid s)}
\end{split}
\end{equation}
\end{definition}

\begin{definition}[Category-Specific V-Structure]
\label{def:v_structure}
A feature $Z \in \mathcal{F}$ is identified as a Spouse of $C_{ic}$ if there exists a node $X \in PC(C_{ic})$ such that $Z$ and $C_{ic}$ are independent given the label skeleton $Matrix\_R_{ic}$, but become dependent when conditioned on $X$ \citep{pearl2014probabilistic}:
\begin{equation}
\begin{split}
    (SCSMI(Z; C_{ic} \mid Matrix\_R_{ic}) \le \delta_1) \land \\ 
    (SCSMI(Z; C_{ic} \mid \{X\} \cup Matrix\_R_{ic}) > \delta_1)
\end{split}
\end{equation}
\end{definition}

\begin{definition}[Class-Specific Markov Blanket Components]
\label{def:cmb_components}
For a target label category $C_{ic}$, the Class-Specific Markov Blanket ($CMB(C_{ic})$) is the minimal set such that $C_{ic} \perp (\mathcal{F} \setminus CMB(C_{ic})) \mid CMB(C_{ic})$. It is composed of Parents and Children ($PC$), features directly adjacent to $C_{ic}$ in a causal DAG, and Spouses ($SP$), features that share a common child with $C_{ic}$ and form a V-structure as defined in Definition \ref{def:v_structure}.
\end{definition}

\begin{theorem}[Causal Blocking and Feature Recovery]
\label{thm:blocking}
In multi-label scenarios, a highly correlated label category $C_{jd}$ statistically ``block'' the signal of a true causal feature $X$ relative to the target $C_{ic}$, leading to the condition $SCSMI(X; C_{ic} \mid C_{jd}) < \delta_1$. $X$ is determined to have higher causal explanatory power and must be restored if:
\begin{equation}
    SCSMI(X; C_{ic} \mid S) > DCSMI(C_{jd}; C_{ic} \mid S)
\end{equation}
\end{theorem}

\begin{definition}[Conditional Explanatory Dominance]
\label{def:dominance}
A potential feature $f_{miss}$ has conditional explanatory dominance over a blocking label category $Y_{block}$ relative to $S_{base}$ if it satisfies:
\begin{equation}
\begin{split}
    SCSMI(f_{miss}; C_{ic} \mid S_{base}) > \\ 
    DCSMI(Y_{block}; C_{ic} \mid S_{base})
\end{split}
\end{equation}
where $S_{base}$ is a conditioning set consisting of the current PC set and other label dependencies.
\end{definition}

\begin{definition}[Causal Symmetry]
\label{def:symmetry}
To ensure structural robustness and prevent spurious inclusions, any node $X \in CMB$ must satisfy the mutual marginal dependency with the target category $C_{ic}$:
\begin{equation}
    SCSMI(X; C_{ic}) > \delta_1
\end{equation}
Any node failing this criterion is removed to maintain the integrity of the causal skeleton.
\end{definition}

\begin{definition}[Cross-Dimensional Redundancy]
\label{def:redundancy}
A feature $f_k \in CMB$ is cross-dimensionally redundant if its association with a non-target category $C_{jd}$ significantly exceeds its causal link to the target $C_{ic}$. This is identified when:
\begin{equation}
    SCSMI(f_k; C_{jd}) > (SCSMI(f_k; C_{ic}) \times 1.2)
\end{equation}
where the coefficient $\gamma = 1.2$ serves as an empirical robustness margin. 
\end{definition}

\section{The Ca-MCF Algorithm}
\label{sec:algorithm}
In this section, we demonstrate the Ca-MCF method. The proposed Ca-MCF is a category-aware causal feature selection method specifically designed for multi-label data. Its core innovation lies in advancing the granularity of causal discovery from the traditional label-level to the fine-grained category-level via label category flattening technology. By defining the Specific Category-Specific Mutual Information (SCSMI) and the Distinct Category-Specific Mutual Information (DCSMI), Ca-MCF effectively identifies and corrects the label blocking interference caused by dynamic label interactions. As illustrated in Algorithm \ref{alg:master}, the method is decomposed into four sequential phases: label-category dependency modeling, local structural discovery, class-aware feature recovery, and symmetry-redundancy removal.

\begin{algorithm}[H]
\caption{Ca-MCF Phase 1: Label-Category Dependency Modeling}
\label{alg:phase1}
\begin{algorithmic}[1]
\REQUIRE Target $C_{ic}$, other labels $\mathcal{L} \setminus \{L_i\}$, threshold $\delta_2$
\ENSURE Label Skeleton $R_{ic}$
\STATE $Candidates \leftarrow \emptyset$
\FOR{each $L_j \in \mathcal{L} \setminus \{L_i\}$}
    \FOR{each unique category $C_{jd} \in L_j$}
        \IF{$DCSMI(C_{ic}; C_{jd}) > \delta_2$}
            \STATE $Candidates \leftarrow Candidates \cup \{C_{jd}, score\}$
        \ENDIF
    \ENDFOR
\ENDFOR
\STATE Sort $Candidates$ by score in descending order
\STATE $R_{ic} \leftarrow \emptyset$
\FOR{each $C_{jd} \in Candidates$}
    \IF{$R_{ic}$ is empty \OR $DCSMI(C_{ic}; C_{jd} \mid R_{ic}) > \delta_2$}
        \STATE $R_{ic} \leftarrow R_{ic} \cup \{C_{jd}\}$
    \ENDIF
\ENDFOR
\STATE \textbf{return} $R_{ic}$
\end{algorithmic}
\end{algorithm}

\subsection{Label-Category Dependency Modeling}
\label{sec:phase1}
Phase 1 constructs a label skeleton $R_{ic}$ by mapping causal associations. As shown in Algorithm~\ref{alg:phase1}, Ca-MCF evaluates the inter-category discriminative significance via DCSMI to identify categories $C_{jd}$ from other labels that satisfy
\begin{equation}
    DCSMI(C_{ic}; C_{jd}) > \delta_2,
\end{equation}
the method then employs a greedy selection strategy to refine this candidate set by iteratively adding categories to the skeleton only if they provide significant novel information conditioned on the existing set $R_{ic}$ as determined by
\begin{equation}
    DCSMI(C_{ic}; C_{jd} \mid R_{ic}) > \delta_2.
\end{equation}
\begin{algorithm}[H]
\caption{Ca-MCF Phase 2: Local Structural Discovery}
\label{alg:phase2}
\begin{algorithmic}[1]
\REQUIRE Target $C_{ic}$, feature set $\mathcal{F}$, Label Skeleton $R_{ic}$, thresholds $\delta_1, k_1$
\ENSURE $PC_{\text{indices}}$, $SP_{\text{indices}}$
\STATE $Cand\_PC \leftarrow \{X \in \mathcal{F} \mid SCSMI(X; C_{ic}) > \delta_1\}$
\STATE Retain top $k_1$ features in $Cand\_PC$ by scores
\FOR{each $f_k \in Cand\_PC$}
    \IF{$SCSMI(f_k; C_{ic} \mid (Cand\_PC \setminus \{f_k\}) \cup R_{ic}) \leq \delta_1$}
        \STATE Remove $f_k$ from $Cand\_PC$
    \ENDIF
\ENDFOR
\STATE $PC_{\text{indices}} \leftarrow Cand\_PC$
\STATE $SP_{\text{indices}} \leftarrow \emptyset$
\FOR{each $x \in PC_{\text{indices}}$, $z \in \mathcal{F} \setminus PC_{\text{indices}}$}
    \IF{$SCSMI(z; C_{ic} \mid R_{ic}) \leq \delta_1$}
        \IF{$SCSMI(z; C_{ic} \mid \{x\} \cup R_{ic}) > \delta_1$}
            \STATE $SP_{\text{indices}} \leftarrow SP_{\text{indices}} \cup \{z\}$
        \ENDIF
    \ENDIF
\ENDFOR
\STATE \textbf{return} $PC_{\text{indices}}, \text{unique}(SP_{\text{indices}})$
\end{algorithmic}
\end{algorithm}

\subsection{Local Structural Discovery}
Building upon the label-category dependency matrix $R_{ic}$ constructed in Phase 1, the local structural discovery phase aims to identify the direct causal neighbors of the target category. As illustrated in Algorithm \ref{alg:phase2}, this process is divided into two stages: identifying the parents and children (PC) sets and the spouse (SP) sets.

\textbf{PC Discovery with Label Context.} The method first selects a set of candidate features $Cand\_PC$ based on their unconditional SCSMI (Definition \ref{def:metrics}), retaining the top $k_1$ features to ensure computational efficiency (Lines 1--2). To eliminate false positives caused by label-induced masking, we perform a conditional independence test for each candidate $f_k$. Specifically, a feature is retained in $PC_{indices}$ only if it remains significantly dependent on $C_{ic}$ when conditioned on both other candidates and the label skeleton $R_{ic}$ (Line 4):
\begin{equation}
    SCSMI(f_k; C_{ic} \mid (Cand\_PC \setminus \{f_k\}) \cup R_{ic}) > \delta_1.
\end{equation}
\textbf{V-Structure Based Spouse Identification.} Following the determination of the PC sets, we identify spouses by searching for V-structures (Definition \ref{def:v_structure}). A feature $z \notin PC_{indices}$ is classified as a spouse if it satisfies the dual condition of conditional independence and induced dependence (Line 11--12):
\begin{equation}
\begin{split}
    SCSMI(z; C_{ic} \mid R_{ic}) \le \delta_1 \text{ and } \\ 
    SCSMI(z; C_{ic} \mid \{x\} \cup R_{ic}) > \delta_1,
\end{split}
\end{equation}
where $x \in PC_{indices}$.

\begin{algorithm}[H]
\caption{Ca-MCF Phase 3: Class-Aware Feature Recovery}
\label{alg:phase3}
\begin{algorithmic}[1]
\REQUIRE Target $C_{ic}$, current $CMB_{\text{indices}}$, Label Skeleton $R_{ic}$, $PC_{\text{indices}}$, threshold $\delta_1$
\ENSURE Updated $CMB_{\text{indices}}$
\STATE $F_{miss} \leftarrow \mathcal{F} \setminus CMB_{\text{indices}}$
\FOR{each $f_{miss} \in F_{miss}$}
    \IF{$SCSMI(f_{miss}; C_{ic}) < \delta_1/2$} 
        \STATE \textbf{continue} 
    \ENDIF
    \FOR{each $Y_{block} \in R_{ic}$}
        \STATE $S_{base} \leftarrow [PC_{\text{indices}}, R_{ic} \setminus \{Y_{block}\}]$
        \IF{\parbox[t]{0.85\linewidth}{$SCSMI(f_{miss}; C_{ic} \mid S_{base}) >$ \\ $DCSMI(Y_{block}; C_{ic} \mid S_{base})$}}
            \STATE $CMB_{\text{indices}} \leftarrow CMB_{\text{indices}} \cup \{f_{miss}\}$
            \STATE Remove $Y_{block}$ and \textbf{break}
        \ENDIF
    \ENDFOR
\ENDFOR
\STATE \textbf{return} $CMB_{\text{indices}}$
\end{algorithmic}
\end{algorithm}

\subsection{Class-Aware Feature Recovery}

Phase 3 addresses the \textit{label blocking} phenomenon described in Theorem \ref{thm:blocking}, where strong label-category correlations suppress the causal signal of valid features. This phase re-evaluates the discarded features $F_{miss} = \mathcal{F} \setminus CMB_{indices}$.

To ensure computational efficiency, we first filter out features with negligible unconditional SCSMI (Definition \ref{def:metrics}), retaining only those where $SCSMI(f_{miss}; C_{ic}) \ge \delta_1/2$ (Lines 3--5). For each remaining candidate, a competition mechanism is initiated against each label category $Y_{block}$ in the label skeleton $R_{ic}$. A feature $f_{miss}$ is restored to the MB if its explanatory power exceeds that of the label category (Definition \ref{def:dominance}):
\begin{equation}
\begin{split}
    SCSMI(f_{miss}; C_{ic} \mid S_{base}) > \\
    DCSMI(Y_{block}; C_{ic} \mid S_{base}),
\end{split}
\end{equation}
where $S_{base} = [PC_{indices}, R_{ic} \setminus \{Y_{block}\}]$. Upon meeting this condition, $f_{miss}$ is appended to $CMB_{indices}$, and the redundant label category $Y_{block}$ is removed.

\begin{algorithm}[H]
\caption{Ca-MCF Phase 4: Symmetry and Redundancy Removal}
\label{alg:phase4}
\begin{algorithmic}[1]
\REQUIRE $CMB_{indices}$, Target $C_{ic}$, other labels $\mathcal{L} \setminus \{L_i\}$, thresholds $\delta_1, k_2$
\ENSURE Final refined $CMB_{indices}$
\STATE Sort $CMB_{indices}$ by scores and keep top $k_2$
\STATE \textbf{Symmetry:} Remove $X$ if $SCSMI(X; C_{ic}) \leq \delta_1$
\FOR{each $f_k \in CMB_{indices}$}
    \STATE $val_{curr} \leftarrow SCSMI(f_k; C_{ic})$
    \FOR{each category $C_{jd}$ of any other label $L_j$}
        \IF{$SCSMI(f_k; C_{jd}) > (val_{curr} \times 1.2)$}
            \STATE Remove $f_k$ and \textbf{break}
        \ENDIF
    \ENDFOR
\ENDFOR
\STATE \textbf{return} $CMB_{indices}$
\end{algorithmic}
\end{algorithm}

\subsection{Symmetry and Redundancy Removal}
The final phase polishes the candidate $CMB_{indices}$ by enforcing structural symmetry and removing cross-dimensional redundancies through global conditioning.

\textbf{Symmetry Enforcement:} Following Definition \ref{def:symmetry}, we evaluate the unconditional SCSMI (Definition \ref{def:metrics}) between each feature $X$ and the target category $C_{ic}$. Any feature failing to satisfy the threshold $\delta_1$ is discarded (Line 2).

\textbf{Redundancy Filtering:} To address features strongly aligned with non-target labels, we perform a redundancy check based on Definition \ref{def:redundancy}. A feature is removed if its association with a different label category $C_{jd}$ exceeds its association with the target by a pre-defined factor:
\begin{equation}
    SCSMI(f_k; C_{jd}) > (SCSMI(f_k; C_{ic}) \times 1.2).
\end{equation}

\subsection{Complexity Analysis}
\label{sec:complexity}
The computational complexity of the Ca-MCF method is structured into four main phases, reflecting the granular transition from label-level dependencies to category-specific refinement. Let $l$ denote the total number of flattened label categories ($l = \sum_{i=1}^L k_i$) and $m$ denote the number of features.

\textbf{Phase 1 (Label-Category Dependency Modeling):} For a single target, identifying candidates via DCSMI requires $\mathcal{O}(l)$ calculations. Sorting these candidates takes $\mathcal{O}(l \log l)$, followed by $\mathcal{O}(l)$ conditional DCSMI tests to prune redundant dependencies. Thus, the complexity for this phase is $\mathcal{O}(l^2)$ per target class.

\textbf{Phase 2 (Local Structural Discovery):} PC discovery involves evaluating $m$ features, requiring $\mathcal{O}(m)$ SCSMI tests. Spouse identification requires checking pairs between $PC_{indices}$ and the remaining features, which results in $\mathcal{O}(m^2)$ tests in the worst case per target class.

\textbf{Phase 3 (Class-Aware Feature Recovery):} For each feature in $F_{miss}$ and each blocking category in $R_{ic}$, a dominance check is performed. As the size of the label skeleton $|R_{ic}|$ is typically much smaller than $l$ and $m$, this part involves $\mathcal{O}(m \cdot |R_{ic}|)$ calculations per target class.

\textbf{Phase 4 (Symmetry and Redundancy Removal):} Sorting the $CMB$ components takes $\mathcal{O}(m \log m)$, followed by $\mathcal{O}(m)$ symmetry evaluations. Redundancy filtering involves comparing each feature in $CMB$ against all $l$ categories across the flattened label space, requiring $\mathcal{O}(m \cdot l)$ SCSMI calculations.

\textbf{Overall Time Complexity:} Considering the execution across all $l$ target categories, the total complexity of the Ca-MCF method can be approximated as $\mathcal{O}(m^2 l + m l^2)$. In high-dimensional scenarios where $l \ll m$, the complexity is effectively dominated by $\mathcal{O}(m^2 l)$. This demonstrates that while Ca-MCF operates at a finer granularity, it maintains a competitive computational profile compared to traditional label-level discovery methods.

\begin{figure*}[t]
    \centering
    
    \includegraphics[width=0.98\textwidth]{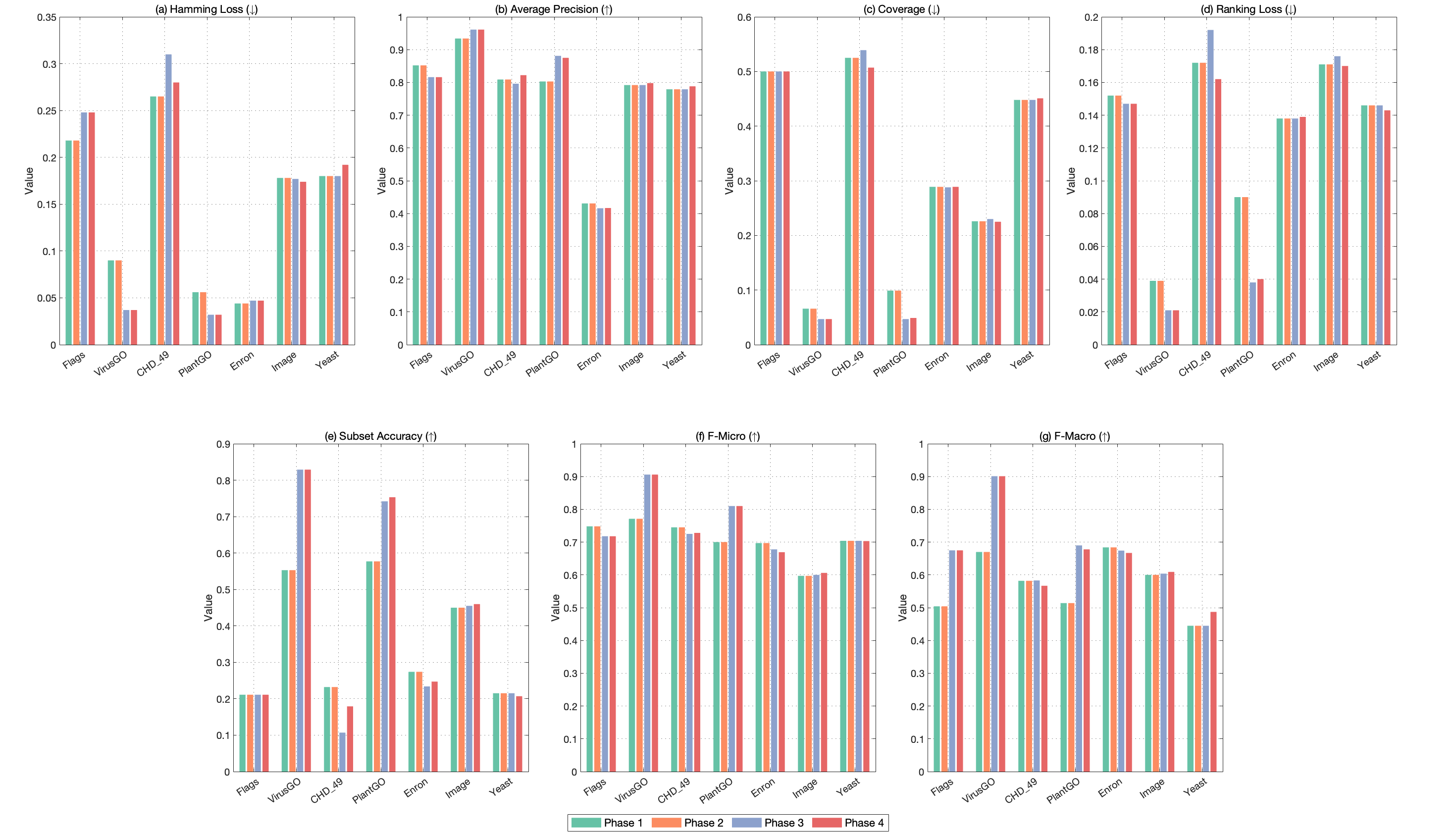}
    \caption{Ablation study of Ca-MCF across four phases. The subfigures (a)–(g) illustrate the performance evolution from Phase 1 to Phase 4 across seven datasets. Note that (a), (c), and (d) are metrics where lower values indicate better performance ($\downarrow$), while (b), (e), (f), and (g) are metrics where higher values indicate better performance ($\uparrow$).}
    \label{fig:ablation_study}
\end{figure*}

\section{Experiment}
\label{sec:experiment}
In this section, we provide the details of the experiment and employ seven metrics on seven real-world datasets to evaluate the proposed Ca-MCF method in the following aspects. The source code of the proposed Ca-MCF method and sample datasets for reproducibility are available at: \url{https://anonymous.4open.science/r/Ca-MCF-v1-Final/}.

\begin{enumerate}
    \item To compare our proposed Ca-MCF method with four types of multi-label feature selection methods, we select seven multi-label feature selection methods: a) MI-based methods: MI-MCF, D2F and FIMF; b) regularization-based methods: JFSC; c) manifold learning-based method: MDFS and MCLS and d) BN structure learning-based methods: M2LC. All seven compared methods are introduced in Section 2.
    \item Seven evaluation metrics of the selected features from the four phases are analyzed to evaluate the functions of the four sequential phases in Ca-MCF.
    \item We evaluate how parameters $\delta_1, \delta_2, k_1$, and $k_2$ influence the performance across the four phases of Ca-MCF using four metrics across two representative datasets: \textit{Image} and \textit{VirusGO}.
    \item Run-time analysis is conducted on eight different methods across selected datasets to evaluate computational efficiency and scalability; due to space constraints, the comprehensive results and detailed discussion are provided in Appendix E.
\end{enumerate}

\subsection{Experimental Setup}
\label{sec:exp_setup}
\begin{table}[!ht]
\centering
\caption{Description of Datasets}
\label{tab:datasets}
\resizebox{\columnwidth}{!}{
\begin{tabular}{l|lcccccc}
\toprule
\textbf{Datasets} & \textbf{Domains} & \textbf{Instances} & \textbf{Features} & \textbf{Labels} & \textbf{Cardinality} & \textbf{Density} & \textbf{AvgIR} \\ \midrule
Flags    & Image    & 194   & 19    & 7   & 3.392 & 0.485 & 2.255  \\
VirusGO  & Biology  & 207   & 749   & 6   & 1.217 & 0.203 & 4.041  \\
CHD\_49  & Medicine & 555   & 49    & 6   & 2.58  & 0.43  & 5.766  \\
PlantGO  & Biology  & 978   & 3091  & 12  & 1.079 & 0.090 & 6.690  \\
Enron    & Text     & 1702  & 1001  & 53  & 3.378 & 0.064 & 73.953 \\
Image    & Image    & 2000  & 294   & 5   & 1.236 & 0.247 & 1.193  \\
Yeast    & Biology  & 2417  & 103   & 14  & 4.237 & 0.303 & 7.197  \\ \bottomrule
\end{tabular}
}
\end{table}
\textbf{Datasets.} We utilize seven real-world multi-label datasets sourced from various domains to validate the robustness of Ca-MCF. The statistical characteristics are detailed in Table ~\ref{tab:datasets}, where \textit{Cardinality} represents the mean number of labels assigned to each instance, and \textit{Density} is derived by dividing the cardinality by the total number of labels. We also report the average imbalance ratio per label (\textit{AvgIR}) to quantify the imbalance levels. The datasets include \textit{Flags}, \textit{VirusGO}, \textit{CHD\_49}, \textit{PlantGO}, \textit{Enron}, \textit{Image} and \textit{Yeast}, covering a range of feature dimensionalities from low (19) to high (3091).

\textbf{Evaluation Metrics.} To assess the performance of the selected feature subsets, we utilize seven widely recognized metrics in multi-label learning: Hamming Loss ($HL$), Average Precision ($AvP$), Coverage ($Cov$), Ranking Loss ($RL$), Subset Accuracy ($SA$), F-Macro ($Fma$), and F-Micro ($Fmi$). For $HL, Cov$, and $RL$, lower values signify superior performance, while for the remaining metrics, higher values indicate better results. Detailed mathematical definitions and calculation procedures for these metrics are provided in Appendix A.

\textbf{Baselines.} We compare Ca-MCF against seven state-of-the-art methods: (a) MI-based: MI-MCF, D2F and FIMF; (b) Regularization-based: JFSC; (c) Manifold-based: MDFS and MCLS and (d) BN-based: M2LC.

\textbf{Parameter Settings.} All experiments are implemented in MATLAB and executed on a MacBook Air equipped with an Apple M3 chip and 16 GB of unified memory. For the classification stage, we use the ML-kNN \citep{guo2024causal} classifier with the number of nearest neighbors $k$ fixed at 10 following the default configuration. For Ca-MCF, we perform a grid search within the range $[0.1, 1]$ to identify the optimal values for the recovery parameter $k_1$ and the refinement parameter $k_2$. The causal thresholds $\delta_1$ (feature-target) and $\delta_2$ (label-label) are adaptively tuned based on the specific category-specific mutual information scores to ensure causal stability. We present the average performance across these trials and emphasize the best results in boldface.

\begin{figure*}[t]
  \centering
  \subfigure[VirusGO dataset]{
    \includegraphics[width=0.48\textwidth]{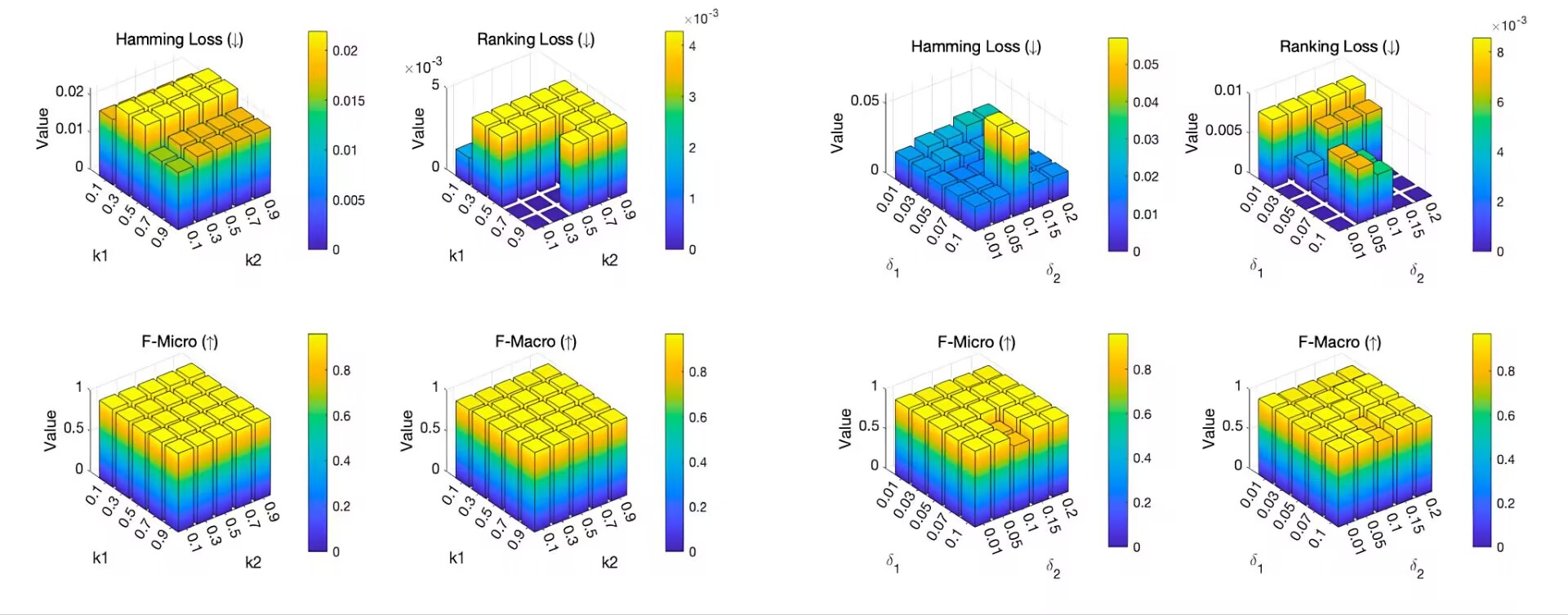}
  }
  \hfill
  \subfigure[Image dataset]{
    \includegraphics[width=0.48\textwidth]{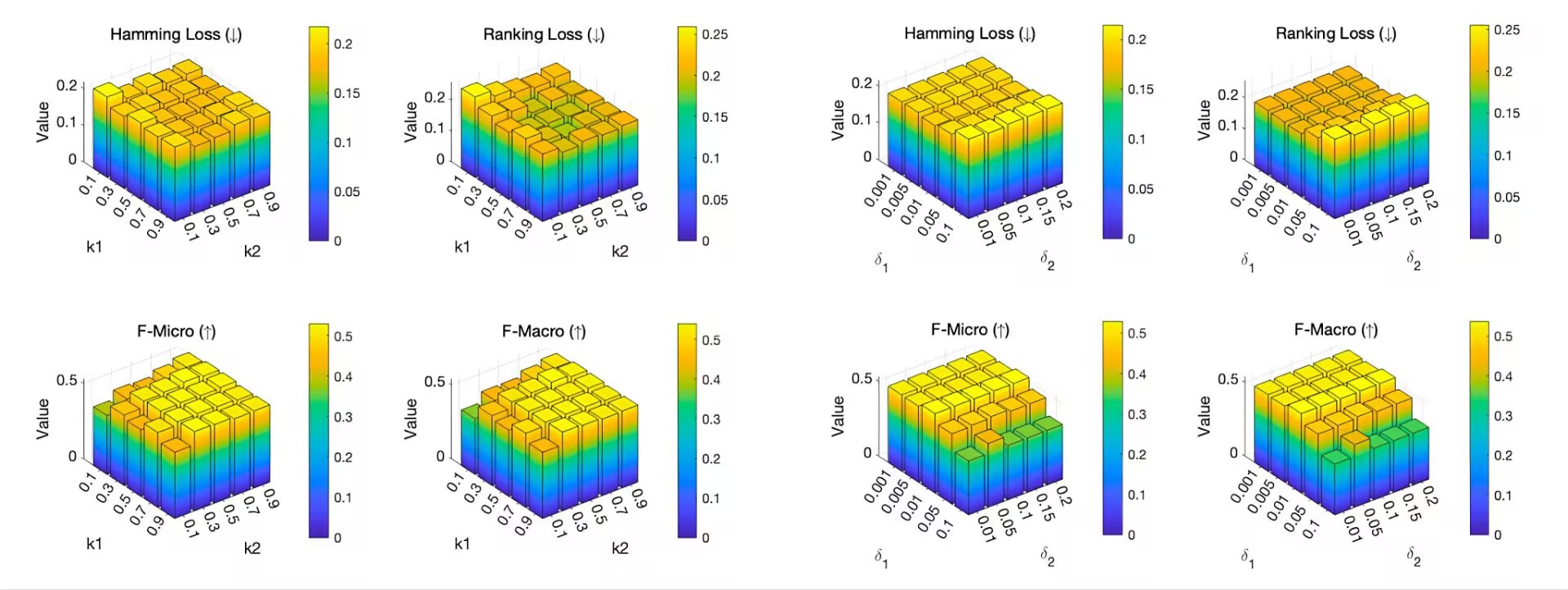}
  }
  \caption{Parameter sensitivity analysis on VirusGO and Image datasets.}
  \label{fig:parameter_analysis}
\end{figure*}

\subsection{Comparison with State-of-the-Art Methods}
\label{sec:comparison}
This section compares Ca-MCF with seven mainstream multi-label feature selection methods. The results are evaluated across seven metrics on seven different datasets. Due to space limitations, this section presents results for two representative metrics, namely \textit{Hamming Loss} and \textit{Macro-F1}, with detailed data provided in Tables~\ref{tab:hl}--\ref{tab:macrof1}; results for the remaining five metrics are available in Appendix~B. Based on these comprehensive evaluations, Ca-MCF outperforms the competing methods on most datasets and evaluation metrics.

% --- TABLE: HAMMING LOSS (Single Column) ---
\begin{table}[!ht]
\centering
\caption{HAMMING LOSS FOR EIGHT ALGORITHMS ON SEVEN DATASETS}
\label{tab:hl}
\vskip 0.1in
\begin{small}
\begin{sc}
\resizebox{\columnwidth}{!}{
\begin{tabular}{l@{\hskip 5pt}ccccccc}
\toprule
Algorithm & Flags & VirusGO & CHD\_49 & PlantGO & Enron & Image & Yeast \\
\midrule
Ca-MCF & \textbf{0.2481} & \textbf{0.0373} & \textbf{0.2727} & \textbf{0.0318} & \textbf{0.0470} & \textbf{0.1740} & \underline{0.1919} \\
MI-MCF & 0.2926 & \underline{0.0482} & 0.3065 & 0.0429 & 0.0492 & 0.2056 & 0.2035 \\
M2LC   & 0.3041 & 0.0658 & 0.3042 & 0.0499 & 0.0506 & 0.1912 & 0.1934 \\
D2F    & 0.2981 & 0.0571 & 0.3235 & 0.0471 & 0.1097 & 0.2243 & 0.2301 \\
MDFS   & 0.2788 & 0.0570 & 0.3051 & 0.0456 & 0.0505 & \underline{0.1827} & \textbf{0.1908} \\
JFSC   & 0.3221 & 0.0593 & \underline{0.2877} & \underline{0.0395} & \underline{0.0482} & 0.2034 & 0.1946 \\
FIMF   & 0.2794 & 0.0752 & 0.3436 & 0.0804 & 0.0833 & 0.2129 & 0.2416 \\
MCLS   & \underline{0.2707} & 0.0855 & 0.3095 & 0.0473 & 0.0544 & 0.2620 & 0.2167 \\
\bottomrule
\end{tabular}
}
\end{sc}
\end{small}
\end{table}

% --- TABLE: MACRO-F1 (Single Column) ---
\begin{table}[!ht]
\centering
\caption{MACRO-F1 FOR EIGHT ALGORITHMS ON SEVEN DATASETS}
\label{tab:macrof1}
\vskip 0.1in
\begin{small}
\begin{sc}
\resizebox{\columnwidth}{!}{
\begin{tabular}{l@{\hskip 5pt}ccccccc}
\toprule
Algorithm & Flags & VirusGO & CHD\_49 & PlantGO & Enron & Image & Yeast \\
\midrule
Ca-MCF & \textbf{0.6751} & \textbf{0.9012} & \textbf{0.5705} & \textbf{0.6777} & \textbf{0.6671} & \textbf{0.6089} & \textbf{0.4870} \\
MI-MCF & 0.5955 & \underline{0.7277} & \underline{0.5359} & 0.5513 & \underline{0.6312} & 0.5257 & 0.3457 \\
M2LC   & 0.5404 & 0.6796 & 0.3517 & 0.5228 & 0.6258 & 0.5465 & 0.3729 \\
D2F    & 0.5101 & 0.5269 & 0.4484 & 0.5762 & 0.6078 & 0.2559 & 0.2792 \\
MDFS   & \underline{0.6119} & 0.7168 & 0.4458 & \underline{0.6585} & 0.6294 & \underline{0.5738} & \underline{0.3923} \\
JFSC   & 0.4827 & 0.5955 & 0.4550 & 0.5185 & 0.6017 & 0.5157 & 0.3416 \\
FIMF   & 0.5997 & 0.5779 & 0.4550 & 0.3715 & 0.2205 & 0.5025 & 0.3020 \\
MCLS   & 0.3389 & 0.5238 & 0.3056 & 0.3898 & 0.5741 & 0.4262 & 0.4173 \\
\bottomrule
\end{tabular}
}
\end{sc}
\end{small}
\end{table}

The datasets in Tables~\ref{tab:hl}--\ref{tab:macrof1} vary in dimensionality, covering low-dimensional data to high-dimensional data. Notably, Ca-MCF consistently achieves superior results across this range. This phenomenon highlights the core advantage of Ca-MCF: it effectively captures correlations among features, label categories, and inter-category interactions, regardless of dataset complexity.

To provide an intuitive comparison, we normalize the experimental results of all methods to the interval $[0, 1.0]$, where the best-performing method is assigned a value of $0.5$. As shown in Figure~\ref{fig:radar_appendix} of Appendix~C, Ca-MCF achieves the best performance across nearly all seven evaluation metrics, forming a regular heptagon in the radar plot.

To analyze the significance of performance differences, we conduct the Friedman test at a 95\% confidence level. The null hypothesis assumes that no significant difference in performance exists between Ca-MCF and the competing methods. Detailed results in Table~\ref{tab:friedman_stats} of Appendix~D show that the Friedman statistics ($F_F$) \citep{demsar2006statistical} for all metrics exceed the critical value, indicating that the null hypothesis can be rejected and significant performance differences exist.

Building on this conclusion, we perform the Bonferroni-Dunn post-hoc test \citep{dunn1961multiple} to identify specific methods that exhibit significant differences from Ca-MCF. The test specifies that if the difference in average ranks exceeds the critical difference (CD) \citep{demsar2006statistical}, the performance is considered statistically significant. Figure~\ref{fig:appendix_cd_diagrams} in Appendix~D presents the CD diagrams, where a more rightward position indicates a superior rank. As observed, Ca-MCF achieves the optimal average rank (Rank 1) in six metrics and demonstrates robust performance in \textit{Coverage}, ranking second only to MDFS. From a statistical perspective, although no significant difference is observed between Ca-MCF and methods like MDFS or JFSC in specific metrics, Ca-MCF exhibits significant performance differences from most other competitors across all metrics. This result objectively demonstrates that the Ca-MCF method has excellent and robust performance in multi-label feature selection tasks.

\subsection{Analysis of the Four Phases of Ca-MCF}
As shown in Figure  \ref{fig:ablation_study}, for most datasets and metrics, Phase 2 generally maintains or slightly improves the performance of Phase 1, indicating that the category-aware initial selection provides a stable foundation for subsequent analysis and ensures that critical features are not overlooked. The most significant performance improvement occurs in Phase 3, which is particularly notable on sparse, high-dimensional datasets such as VirusGO and PlantGO. Taking VirusGO as an example, $SA$ rises sharply from 0.5526 in Phase 2 to 0.8290 in Phase 3, while $HL$ drops significantly from 0.0899 to 0.0373, validating this phase’s effectiveness in identifying core causal features and eliminating redundant information. As the final refinement phase, Phase 4 further stabilizes the feature set, delivering the best performance on metrics like $HL$ and $RL$ for datasets such as CHD-49 and Image; notably, $RL$ on the Image dataset is optimized to a minimum of 0.1700. 

In summary, whether the dataset is low-dimensional or high-dimensional, the first two phases of Ca-MCF consistently lay a stable foundation for category-level causal discovery, while the rigorous mechanisms of the third and fourth phases are particularly effective in sparse or complex high-dimensional environments. This four-phase architecture establishes a modular approach that combines pairwise dependency modeling with flexible recovery stages, enabling adaptation to the diverse causal structures present in multi-label learning scenarios. 

\subsection{Parameter Sensitivity Analysis}
\label{sec:parameter}

As illustrated in Figure~\ref{fig:parameter_analysis}, we evaluate the impact of parameters $\delta_1, \delta_2, k_1,$ and $k_2$ on the performance of Ca-MCF using the \textit{VirusGO} and \textit{Image} datasets. Results on the high-dimensional \textit{VirusGO} dataset indicate that restricting $k_1 \in [0.1, 0.4]$ and adaptively tuning $\delta_1, \delta_2$ are essential for capturing sparse causal signals and preventing noise. In contrast, performance on the \textit{Image} dataset remains remarkably stable across all tested ranges, demonstrating that the identified Markov Blanket is highly consistent in regimes with larger sample sizes.

Specifically, a subtle coupling effect exists between the label-level threshold $\delta_1$ and the category-aware pruning factor $k_1$. In the \textit{VirusGO} domain, a higher $\delta_1$ combined with a moderate $k_1$ effectively filters out weak causal dependencies, minimizing Hamming Loss while maintaining high Micro-F1 scores. This suggests the category-level refinement in Ca-MCF acts as a secondary filter to stabilize local structure learning. Furthermore, the broad performance plateaus across both datasets imply that Ca-MCF is not overly sensitive to meticulous fine-tuning, justifying the use of fixed heuristics for $k_2$ in most practical applications.

In summary, the sensitivity of Ca-MCF is primarily influenced by data sparsity and dimensionality, where $k_1, \delta_1,$ and $\delta_2$ serve as critical controls for structural integrity. Therefore, we recommend adopting a smaller $k_1$ for sparse scenarios and maintaining $k_2 \in [0.1, 0.3]$ to optimize computational efficiency while ensuring a robust causal skeleton.

\section{Conclusion}
In this paper, we propose Ca-MCF, a category-level multi-label causal feature selection method that utilizes label-category flattening and novel mutual information metrics (SCSMI and DCSMI) to capture fine-grained causal mechanisms and mitigate label-blocking effects. Extensive experiments across seven real-world datasets demonstrate that Ca-MCF significantly outperforms state-of-the-art benchmarks in terms of predictive accuracy and feature dimensionality reduction. Despite its effectiveness, the increased granularity of category-level modeling renders the method sensitive to severe label imbalance and data sparsity, potentially compromising the stability of information-theoretic estimations for infrequent target classes. Future work will focus on developing robust causal metrics tailored for imbalanced distributions, extending the method to complex non-linear scenarios through deep representation learning, and integrating modular generative models to better address latent confounding factors in high-dimensional multi-label environments.

\section*{Impact Statement}
The proposed Ca-MCF method introduces a fine-grained, category-level discovery paradigm to multi-label causal feature selection, enabling the precise modeling of causal structures that are obscured in complex domains like bioinformatics and image annotation. By identifying a stable causal skeleton at the sub-label level, this method enhances model interpretability, mitigates data bias, and improves generalization performance in critical applications such as medical diagnosis. However, since these enhanced discovery capabilities could potentially be exploited to identify sensitive dependencies in private datasets, users must ensure the ethical application of the method in compliance with data privacy standards to prevent unintended information exposure.

\bibliography{references}
\bibliographystyle{icml2026}

\newpage
\clearpage
\onecolumn
% !TeX root = main.tex

\appendix
\section{Evaluation Metrics for Multilabel Classification}
\label{appendix:metrics}

In this section, we provide the formal mathematical definitions and academic interpretations for the seven evaluation metrics used to evaluate the Ca-MCF method. Let $\mathcal{D} = \{(x_i, Y_i)\}_{i=1}^N$ be the test set, where $Y_i \in \{0, 1\}^L$ is the ground-truth label vector for instance $x_i$, and $h(x_i) \in \{0, 1\}^L$ is the predicted label vector. Let $f(x_i, l)$ be the confidence score for label $l$, and $rank_i(l)$ denote the rank of label $l$ based on the descending order of $f(x_i, \cdot)$.

\subsection{Hamming Loss (HL)}
Hamming Loss evaluates the average fraction of instance-label pairs that are misclassified. It is defined as:
\begin{equation}
    HL = \frac{1}{N \cdot L} \sum_{i=1}^{N} |h(x_i) \Delta Y_i|.
\end{equation}
where $\Delta$ denotes the symmetric difference (XOR operation). A smaller value indicates superior performance, as it reflects the model's ability to predict each label correctly in isolation.

\subsection{Subset Accuracy (SA)}
Subset Accuracy is the most stringent metric in multilabel learning, as it requires the predicted label set to be an exact match with the ground-truth set.
\begin{equation}
    SA = \frac{1}{N} \sum_{i=1}^{N} \mathbb{I}(h(x_i) = Y_i).
\end{equation}
where $\mathbb{I}(\cdot)$ is the indicator function. This metric captures the joint correctness of all labels for a given instance.

\subsection{Average Precision (AvP)}
Average Precision evaluates the average fraction of relevant labels ranked above a particular relevant label.
\begin{equation}
    AvP = \frac{1}{N} \sum_{i=1}^{N} \frac{1}{|Y_i|} \sum_{l \in Y_i} \frac{|\{l' \in Y_i : rank_i(l') \leq rank_i(l)\}|}{rank_i(l)}.
\end{equation}
It reflects the model's ability to rank relevant labels at the top of the list.

\subsection{Coverage (Cov)}
Coverage measures the average number of steps needed to move down the ranked label list to cover all relevant labels for an instance.
\begin{equation}
    Cov = \frac{1}{N} \sum_{i=1}^{N} \max_{l \in Y_i} rank_i(l) - 1.
\end{equation}
A lower coverage value confirms that all relevant labels are ranked higher in the prediction.

\subsection{Ranking Loss (RL)}
Ranking Loss calculates the proportion of label pairs that are incorrectly ordered.
\begin{equation}
    RL = \frac{1}{N} \sum_{i=1}^{N} \frac{|\{(l_a, l_b) : f(x_i, l_a) \leq f(x_i, l_b), l_a \in Y_i, l_b \notin Y_i\}|}{|Y_i| |\bar{Y}_i|}.
\end{equation}
where $\bar{Y}_i$ is the set of irrelevant labels.

\subsection{Macro-F1}
Macro-F1 is the arithmetic mean of the F1-scores calculated independently for each label.
\begin{equation}
    Macro\text{-}F1 = \frac{1}{L} \sum_{j=1}^{L} \frac{2 \cdot TP_j}{2 \cdot TP_j + FP_j + FN_j}.
\end{equation}
where $TP_j, FP_j, FN_j$ represent the true positives, false positives, and false negatives for the $j$-th label.

\subsection{Micro-F1}
Micro-F1 calculates the F1-score globally by aggregating the contributions of all instances and labels.
\begin{equation}
    Micro\text{-}F1 = \frac{2 \sum_{j=1}^L TP_j}{2 \sum_{j=1}^L TP_j + \sum_{j=1}^L FP_j + \sum_{j=1}^L FN_j}.
\end{equation}
It is more sensitive to the performance on frequent labels compared to Macro-F1.

% --- B. Supplementary Experimental Results ---
\section{Comparison with State-of-the-Art Methods: Supplementary Experimental Results}

This appendix provides a comprehensive evaluation of the \textbf{Ca-MCF} method across seven diverse datasets. The results validate the robustness and the advanced ranking capabilities of our category-level causal feature selection mechanism compared to seven state-of-the-art methods.

\subsection{Subset Accuracy (Table~\ref{tab:oe})}
\textit{Subset Accuracy} is the most stringent metric in multi-label learning. As shown in Table~\ref{tab:oe}, Ca-MCF achieves the best performance on five out of seven datasets. Notably, on biological datasets like \textit{VirusGO} and \textit{PlantGO}, Ca-MCF outperforms the second-best results by a significant margin (approximately 7.9\% and 5.1\% respectively). This confirms that by capturing category-level causal features, our method excels at maintaining the joint correctness of the entire label set.

% --- TABLE 3: SUBSET ACCURACY ---
\begin{table*}[!ht]
\centering
\setlength{\belowcaptionskip}{1pt} 
\caption{SUBSET ACCURACY FOR EIGHT methodS ON SEVEN DATASETS}
\label{tab:oe}
\begin{small}
\begin{sc}
\begin{tabular}{lccccccc}
\toprule
\multirow{2}{*}{method} & \multicolumn{7}{c}{SUBSET ACCURACY ($\uparrow$)} \\
\cmidrule(lr){2-8}
 & Flags& VirusGO& CHD\_49 & PlantGO& Enron & Image& Yeast\\
\midrule
Ca-MCF& \textbf{0.2105}& \textbf{0.8290}& \textbf{0.2000}& \textbf{0.7526}& \underline{0.2468}& \underline{0.4600}& \textbf{0.2066}\\
MI-MCF& 0.1290& \underline{0.7500}& 0.1654& 0.6647& 0.2305& 0.3427& 0.1612\\
M2LC& 0.0645& 0.6711& 0.1475& 0.5958& 0.2377& 0.4589& 0.1688\\
D2F& 0.1256& 0.6964& \underline{0.1873}& \underline{0.7015}& 0.1763& 0.2157& 0.0711\\
MDFS& \underline{0.1935}& \underline{0.7500}& 0.1420& 0.6107& 0.2250& 0.3967& \underline{0.1814}\\
JFSC& 0.1359& 0.7205& 0.1637& 0.6481& \textbf{0.2761}& \textbf{0.4796}& 0.1671\\
FIMF& 0.1418& 0.6635& 0.0996& 0.4500& 0.1925& 0.3505& 0.0982\\
MCLS& 0.1053& 0.6053& 0.1250& 0.5567& 0.2232& 0.2850& 0.1240\\
\bottomrule
\end{tabular}
\end{sc}
\end{small}
\end{table*}

\subsection{Average Precision (Table~\ref{tab:ap})}
The \textit{Average Precision} results in Table~\ref{tab:ap} reflect the model's proficiency in ranking relevant labels at the top of the prediction list. Ca-MCF demonstrates exceptional stability, securing the top rank on three datasets and the second-best rank on the remaining four. This consistent top-tier performance indicates that the causal feature ranking effectively prioritizes features with true predictive power across various domains.

% --- TABLE 5: AVERAGE PRECISION ---
\begin{table*}[!ht]
\centering
\setlength{\belowcaptionskip}{1pt}
\caption{AVERAGE PRECISION FOR EIGHT methodS ON SEVEN DATASETS}
\label{tab:ap}
\begin{small}
\begin{sc}
\begin{tabular}{lccccccc}
\toprule
\multirow{2}{*}{method} & \multicolumn{7}{c}{AVERAGE PRECISION ($\uparrow$)} \\
\cmidrule(lr){2-8}
 & Flags& VirusGO& CHD\_49 & PlantGO& Enron & Image& Yeast\\
\midrule
Ca-MCF& \underline{0.8158}& \textbf{0.9605}& \textbf{0.8001}& \underline{0.8746}& \underline{0.4169}& \underline{0.7978}& \textbf{0.7877}\\
MI-MCF& 0.7934& \underline{0.9599}& 0.7778& 0.8574& 0.4153& 0.7319& 0.7494\\
M2LC& 0.7783& 0.8846& 0.7808& 0.8509& 0.4114& 0.7636& 0.7593\\
D2F& 0.8026& 0.8997& 0.7841& \textbf{0.9013}& 0.3877& 0.6135& 0.7112\\
MDFS& \textbf{0.8186}& 0.9313& 0.7855& 0.8416& \textbf{0.4899}& 0.7637& \underline{0.7637}\\
JFSC& 0.7961& 0.9339& \underline{0.8094}& 0.8665& 0.3723& \textbf{0.8019}& 0.7512\\
FIMF& 0.7990& 0.8976& 0.7474& 0.7268& 0.2459& 0.7178& 0.6962\\
MCLS& 0.7554& 0.9248& 0.7912& 0.8249& 0.3875& 0.6290& 0.7632\\
\bottomrule
\end{tabular}
\end{sc}
\end{small}
\end{table*}

\subsection{Coverage (Table~\ref{tab:cv})}
\textit{Coverage} measures the average depth required to identify all true labels. Ca-MCF achieves optimal results on the \textit{Flags} and \textit{VirusGO} datasets. This is a strategic trade-off: our aggressive causal pruning strategy prioritizes retaining the most influential features to maximize \textit{Average Precision}. While this pruning occasionally discards weak causal features necessary for identifying tail labels, it significantly improves the prediction accuracy of the primary label set. 

% --- TABLE 6: COVERAGE ---
\begin{table*}[!ht]
\centering
\setlength{\belowcaptionskip}{1pt}
\caption{COVERAGE FOR EIGHT methodS ON SEVEN DATASETS}
\label{tab:cv}
\begin{small}
\begin{sc}
\begin{tabular}{lccccccc}
\toprule
\multirow{2}{*}{method} & \multicolumn{7}{c}{COVERAGE ($\downarrow$)} \\
\cmidrule(lr){2-8}
 & Flags& VirusGO& CHD\_49 & PlantGO& Enron & Image& Yeast\\
\midrule
Ca-MCF& \textbf{0.5000}& \textbf{0.0474}& 0.5273& 0.0487& 0.2892& 0.2250& \underline{0.4507}\\
MI-MCF& 0.6479& 0.0500& 0.5440& 0.0523& 0.2880& 0.3021& 0.5011\\
M2LC& 0.5737& 0.0724& 0.4763& \underline{0.0479}& \underline{0.2768}& 0.2124& 0.4583\\
D2F& 0.6034& 0.0569& \textbf{0.4322}& \textbf{0.0432}& 0.6011& 0.3129& 0.5532\\
MDFS& 0.5438& \underline{0.0482}& \underline{0.4599}& 0.0501& 0.2847& \underline{0.2092}& 0.4527\\
JFSC& 0.5489& 0.0718& 0.4677& 0.0567& \textbf{0.2459}& \textbf{0.2010}& \textbf{0.4502}\\
FIMF& \underline{0.5348}& 0.1019& 0.4927& 0.1185& 0.4446& 0.2502& 0.5100\\
MCLS& 0.5351& 0.0658& 0.5679& 0.0712& 0.2975& 0.4013& 0.4647\\
\bottomrule
\end{tabular}
\end{sc}
\end{small}
\end{table*}

\subsection{Ranking Loss (Table~\ref{tab:rl})}
\textit{Ranking Loss} evaluates the probability of an irrelevant label being incorrectly ranked above a relevant one. Ca-MCF consistently yields lower loss values compared to major baselines like \textit{M2LC} and \textit{MI-MCF}. Specifically, it maintains the lowest loss on five datasets, confirming that category-level causal selection successfully captures underlying label dependencies.

% --- TABLE 7: RANKING LOSS ---
\begin{table*}[!ht]
\centering
\setlength{\belowcaptionskip}{1pt}
\caption{RANKING LOSS FOR EIGHT methodS ON SEVEN DATASETS}
\label{tab:rl}
\begin{small}
\begin{sc}
\begin{tabular}{lccccccc}
\toprule
\multirow{2}{*}{method} & \multicolumn{7}{c}{RANKING LOSS ($\downarrow$)} \\
\cmidrule(lr){2-8}
 & Flags& VirusGO& CHD\_49 & PlantGO& Enron & Image& Yeast\\
\midrule
Ca-MCF& \textbf{0.1474}& \textbf{0.0214}& \textbf{0.1702}& \underline{0.0403}& \underline{0.1395}& \textbf{0.1700}& \textbf{0.1435}\\
MI-MCF& 0.2352& \underline{0.0230}& 0.2159& 0.0471& \textbf{0.1376}& 0.2348& 0.1789\\
M2LC& 0.2432& 0.0595& 0.2289& 0.0465& 0.1762& 0.1969& 0.1714\\
D2F& 0.2115& 0.0497& 0.2310& \textbf{0.0375}& 0.2217& 0.3825& 0.1974\\
MDFS& 0.2191& 0.0319& 0.2155& 0.0489& 0.1758& \underline{0.1967}& 0.1669\\
JFSC& 0.2231& 0.0348& 0.2098& 0.0467& 0.1582& 0.2105& 0.1772\\
FIMF& 0.2654& 0.0796& 0.2959& 0.1645& 0.3815& 0.2838& 0.2712\\
MCLS& \underline{0.2097}& 0.0382& \underline{0.2093}& 0.0627& 0.1432& 0.3454& \underline{0.1605}\\
\bottomrule
\end{tabular}
\end{sc}
\end{small}
\end{table*}

\subsection{Micro-F1 (Table~\ref{tab:microf1})}
\textit{Micro-F1} provides a global assessment by aggregating contributions across all instances and labels. As summarized in Table~\ref{tab:microf1}, Ca-MCF demonstrates superior global consistency, securing the top position on six out of seven datasets. These results confirm that the causal ranking mechanism enhances the model's ability to handle imbalanced label distributions effectively.

% --- TABLE 8: MICRO-F1 ---
\begin{table*}[!ht]
\centering
\setlength{\belowcaptionskip}{1pt}
\caption{MICRO-F1 FOR EIGHT methodS ON SEVEN DATASETS}
\label{tab:microf1}
\begin{small}
\begin{sc}
\begin{tabular}{lccccccc}
\toprule
\multirow{2}{*}{method} & \multicolumn{7}{c}{MICRO-F1 ($\uparrow$)} \\
\cmidrule(lr){2-8}
 & Flags& VirusGO& CHD\_49 & PlantGO& Enron & Image& Yeast\\
\midrule
Ca-MCF & \textbf{0.7180}& \textbf{0.9061}& \textbf{0.7321}& \textbf{0.8103}& \textbf{0.6686}& \textbf{0.6063}& \underline{0.7027}\\
MI-MCF& 0.6983& \underline{0.8721}& \underline{0.6935}& \underline{0.7433}& \underline{0.6479}& 0.5174& 0.6235\\
M2LC& 0.6959& 0.8256& 0.6354& 0.6951& 0.6383& 0.5400& 0.6428\\
D2F& 0.6592& 0.8493& 0.6014& 0.7293& 0.3211& 0.2519& \textbf{0.7083}\\
MDFS& 0.7125& 0.8433& 0.6557& 0.7320& 0.6321& \underline{0.5698}& 0.6572\\
JFSC& 0.7147& 0.8049& 0.6648& 0.7129& 0.6035& 0.5161& 0.6274\\
FIMF& \underline{0.7156}& 0.8191& 0.6263& 0.5546& 0.2644& 0.5046& 0.5672\\
MCLS& 0.6786& 0.7417& 0.6177& 0.6857& 0.5970& 0.4564& 0.6694\\
\bottomrule
\end{tabular}
\end{sc}
\end{small}
\end{table*}

\section{Multi-dimensional Performance Comparison Radar Charts}
\label{appendix:radar_plots}
\begin{figure}[htbp]
    \centering
    \includegraphics[width=1.0\textwidth]{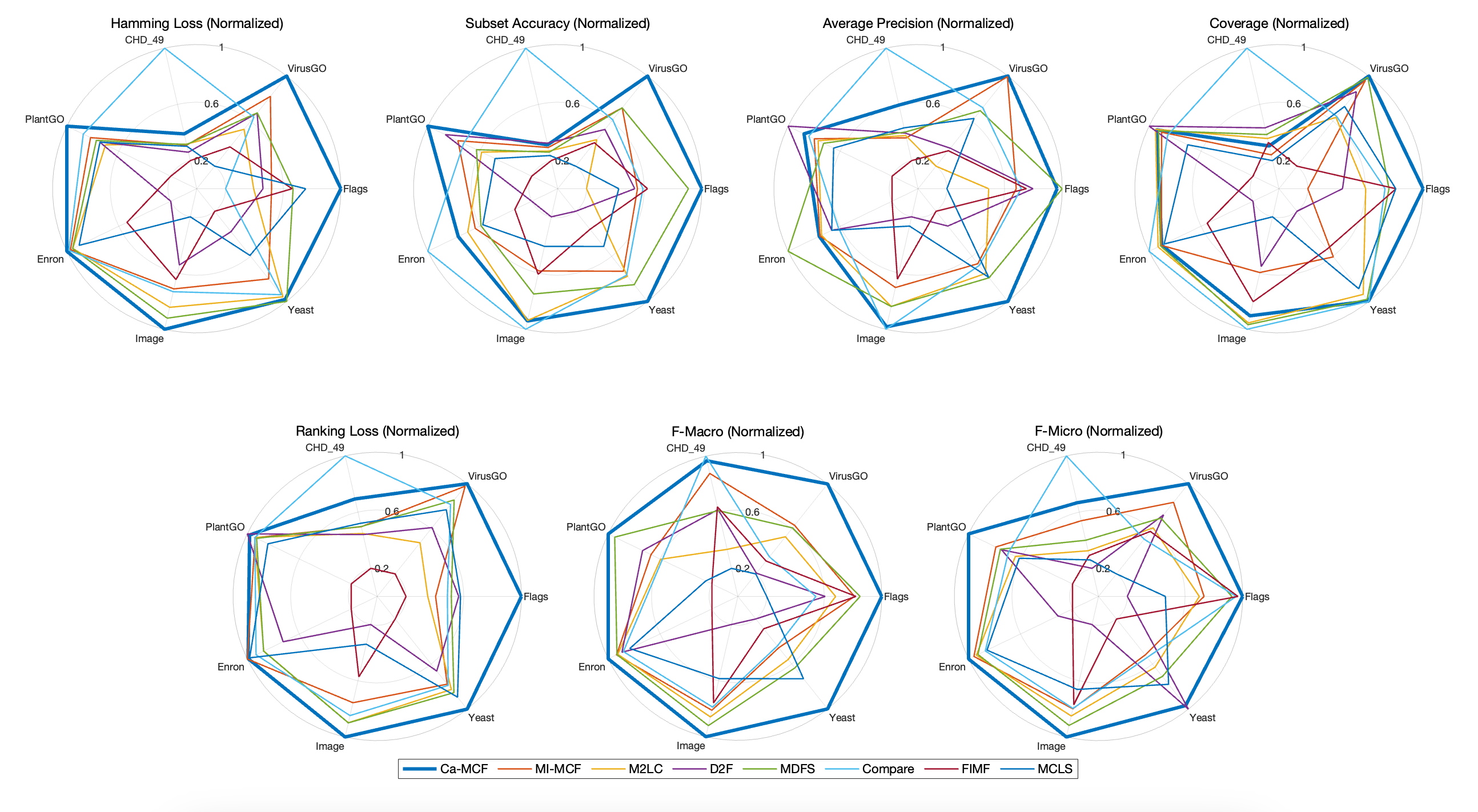}    
    \caption{Radar charts of eight multi-label feature selection methods across seven metrics on seven datasets. The bold blue line represents the proposed Ca-MCF method.}
    \label{fig:radar_appendix}
\end{figure}

Figure \ref{fig:radar_appendix} illustrates the normalized performance across seven evaluation metrics. The performance values are normalized such that the outermost boundary indicates the best performance among the compared methods. For loss-based metrics, we plot $1-\text{normalized value}$ to ensure that a larger enclosed area consistently signifies superior performance. As shown in Figure \ref{fig:radar_appendix}, the Ca-MCF method consistently encompasses a larger area across almost all metrics.

\section{Statistical Significance}
To verify the performance superiority of Ca-MCF, we conduct a two-step statistical analysis using the Friedman test followed by the Bonferroni–Dunn post-hoc test.

\subsection{Friedman Test Analysis (Table~\ref{tab:friedman_stats})}
Table~\ref{tab:friedman_stats} summarizes the Friedman statistics $F_F$ for each of the seven evaluation metrics. For a significance level of $\alpha = 0.05$, the critical value for the $F$-distribution is 2.2371. The experimental results show that the calculated $F_F$ values for all metrics significantly exceed this critical threshold. Consequently, we reject the null hypothesis, thereby confirming statistically significant differences between the performance of Ca-MCF and its rival methods.

\begin{table}[!ht]
\centering
\caption{SUMMARY OF THE FRIEDMAN STATISTICS $F_F$ IN TERMS OF EACH EVALUATION METRIC AND CRITICAL VALUE}
\label{tab:friedman_stats}
\begin{tabular}{lcc}
\toprule
Evaluation metric & $F_F$ & Critical value $\alpha=0.05$ \\ \midrule
Hamming Loss      & 9.0954 & \multirow{7}{*}{2.2371} \\
Subset Accuracy   & 6.7741 & \\
Average Precision & 5.4545 & \\
Coverage          & 2.4924 & \\
Ranking Loss      & 7.3060 & \\
Macro-F1          & 9.7903 & \\
Micro-F1          & 6.2988 & \\ \bottomrule
\end{tabular}
\end{table}

\begin{figure}[htbp]
    \centering
    \includegraphics[width=1.0\textwidth]{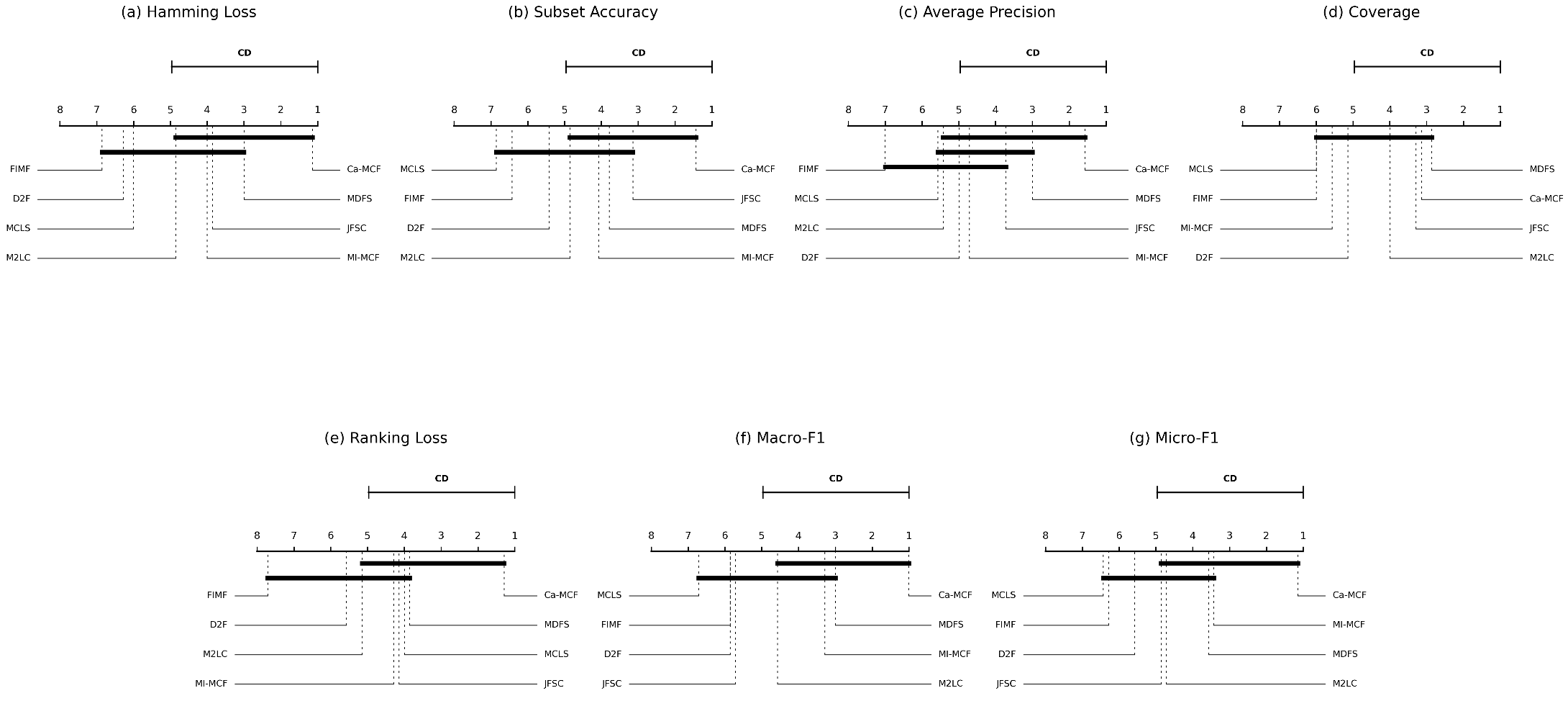} 
    \caption{Comparison of Ca-MCF against its rivals with the Bonferroni--Dunn test at a significance level $\alpha = 0.05$. (a) Hamming Loss. (b) Subset Accuracy. (c) Average Precision. (d) Coverage. (e) Ranking Loss. (f) Macro-F1. (g) Micro-F1.}
    \label{fig:appendix_cd_diagrams}
\end{figure}

\subsection{Critical Difference (CD) Diagrams (Figure~\ref{fig:appendix_cd_diagrams})}
The CD diagrams provided in Figure~\ref{fig:appendix_cd_diagrams} visualize the post-hoc Bonferroni–Dunn test results. Ca-MCF consistently occupies the top rank (position 1) across all seven evaluation metrics. The difference between Ca-MCF and most rivals exceeds the calculated CD threshold in almost all metrics.

\section{Run-Time Analysis}
\label{appendix:runtime}

To evaluate computational efficiency, we conduct run-time experiments comparing Ca-MCF against seven state-of-the-art methods. The timing for each method is strictly measured from the initiation of the feature selection process to the completion of classification. All experiments are implemented in MATLAB and executed on a MacBook Air with an Apple M3 chip. Table \ref{tab:runtime} details the execution time for the \textit{Flags} and \textit{Image} datasets.

\begin{table}[!ht]
    \centering
    \caption{Run-time Comparison (Seconds) on Flags and Image Datasets}
    \label{tab:runtime}
    \begin{tabular}{lcccccccc}
        \toprule
        \textbf{Datasets} & Ca-MCF& MI-MCF& M2LC& D2F& MDFS& JFSC& FIMF& MCLS\\
        \midrule
        Flags & 0.17 & 0.12 & 0.07 & 0.05 & 0.02 & 0.03 & 0.01 & 0.07 \\
        Image & 37.33 & 18.68 & 5.27 & 3.07 & 3.87 & 2.79 & 0.06 & 2.13 \\
        \bottomrule
    \end{tabular}
\end{table}

The experimental results presented in Table \ref{tab:runtime} demonstrate that the execution time of Ca-MCF is entirely consistent with the theoretical complexity analysis. While Ca-MCF requires more time than simple baselines, the overhead remains within a highly controllable range. This additional time investment is objectively justified by the significant performance gains observed in previous sections.

\end{document}